# Identifying the sentiment styles of YouTube's vloggers

**Bennett Kleinberg**
Department of Psychology
University of Amsterdam

Department of Security
and Crime Science
University College London
b.a.r.kleinberg@uva.nl

**Maximilian Mozes**
Department of
Informatics
Technical University
of Munich
mozes@cs.tum.edu

**Isabelle van der Vegt**
Department of Security and
Crime Science
University College London
isabelle.vegt.17@ucl.ac.uk

## Abstract

Vlogs provide a rich public source of data in a novel setting. This paper examined the continuous sentiment styles employed in 27,333 vlogs using a dynamic intra-textual approach to sentiment analysis. Using unsupervised clustering, we identified seven distinct continuous sentiment trajectories characterized by fluctuations of sentiment throughout a vlog's narrative time. We provide a taxonomy of these seven continuous sentiment styles and found that vlogs whose sentiment builds up towards a positive ending are the most prevalent in our sample. Gender was associated with preferences for different continuous sentiment trajectories. This paper discusses the findings with respect to previous work and concludes with an outlook towards possible uses of the corpus, method and findings of this paper for related areas of research.

## 1 Introduction

Vlogging or video-blogging has become one of the most popular video formats on social media platforms like YouTube. Vlogs have been referred to as 'conversational video-blogs' (Biel and Gatica-Perez, 2010) or 'monologue-like' videos (Aran et al., 2014), and are officially defined in the Cambridge dictionary of English as "a record of your thoughts, opinions or experiences that you film and publish on the internet"[1]. The vast amount of vlogs on YouTube comprises a rich body of visual and textual data, usually covering a vlogger's daily life. The transcripts of vlogs provide researchers with ample opportunity to examine natural language in this young field of communication. Nevertheless, little attention has thus far been paid to investigating the language used in YouTube vlogs.

Much of the literature concerning Youtube vlogs focuses on the visual modality (Aran et al., 2014) or meta-indicators such as views and subscriber counts (Borghol et al., 2012). With the current study, we aim to address this gap in the literature by automatically analyzing the linguistic styles used by YouTube's vloggers. Building on a novel approach to examining the continuous sentiment structure, we seek to shed light on the different temporal trajectories used by vloggers, and, by doing so, we expect to gain a deeper understanding of language use in vlogs.

### 1.1 Previous research on vlogs

Research on YouTube vlogs has thus far mainly focused on metadata indicators and how they impact video popularity. For example, a pronounced "rich-get-richer" effect regarding video popularity is often found. When controlling for the content, previous views are the best predictor of later video popularity (Borghol et al., 2012). In addition, video age also predicts popularity: when the content is similar, early uploaders have an advantage over later uploaders in terms of popularity (Borghol et al., 2012). Furthermore, related video recommendations shown during or after a playing video also serve as an important indicator of the view count for YouTube videos (Zhou et al., 2010).

Another strand of research has examined the visual content of vlogs. Across 2,268 single-person YouTube videos, four clusters of user activity level (e.g., stationary or in motion), editing choices, and video quality were identified (Aran et al., 2014). Videos from a cluster with highly edited, active vlogs received more views than simple conversational videos in which the vlogger is stationary in front of the camera. Concerning audio data, videos in which the vlogger is "talking more, faster, and using few pauses" receive more views on average (Biel and Gatica-Perez, 2010). Similarly, the distance of the vlogger to the camera, looking time, and 'looking-while-speaking' were also found to significantly correlate with view count. It has been suggested that these nonverbal characteristics might be related to specific personality traits that promote effective communication and video popularity (Biel and Gatica-Perez, 2010). Lastly, a smaller line of research examines the linguistic content of YouTube vlogs. For example, a manual inspection of a random sample of 100 vlogs suggests that female vloggers are more likely than male vloggers to vlog about personal matters (Molyneaux et al., 2008). In contrast, male YouTubers focused more on entertainment and technology. Research on the linguistic content of vlogs is scarce, which is why the present paper presents a novel methodology and corpus for examining the text modality. In the next section, the proposed method is explained in conjunction with previous research utilizing intra-textual sentiment analysis.

---
[1] https://dictionary.cambridge.org/dictionary/english/vlog

## 1.2 Continuous sentiment trajectories

With increasing amounts of textual data available online, diverse methods for examining natural language and its linguistic style and content are applied on a large scale. Applications range from the automated detection of fake reviews (Ott et al., 2013), fake news (Thorne et al., 2018; Pérez-Rosas et al., 2018) and lies and truths (Mihalcea and Strapparava, 2009; Kleinberg et al., 2018) to detecting political sentiments (Cambria, 2016).

One important strand of computational linguistics examines the sentiment of texts (i.e., how positive or negative the emotional valence of the content is). Typically, a sentiment score is reported for a piece or section of text (e.g., a sentence or a whole text). However, a recently emerged literature examines shifts in emotional valence throughout a text (hence intra-textual) to assess sentiment over time. This focus on the temporal dimension resembles work on storyline extraction, where the timeline of events is considered, including events that give rise to a 'climax' of a story (Caselli and Vossen, 2016). With intra-textual sentiment analysis, it is also possible to visualize language use leading to a climax in (positive or negative) sentiment. This method has thus far been applied to extract the intra-textual sentiment dynamics in novels (Gao et al., 2016) and was used to identify key narrative moments. Within this same context, Jockers (2015a) analyzed the intra-textual sentiments of over 40.000 novels using hierarchical clustering and observed six shapes, each representing a common sentiment structure in novels (e.g., a 'man-in-hole' structure showcasing a positive-negative-positive sentiment).

A few studies have further examined emotional plot shapes in different bodies of text. Six common emotional arcs were found to underlie 1,327 fictional stories, termed the 'rags to riches' (i.e., a rise in sentiment), 'tragedy' or 'riches to rags' (i.e., a fall in sentiment), 'man in hole' (i.e., fall-rise), 'Icarus' (i.e., rise-fall), 'Cinderella' (i.e., rise-fall-rise), and 'Oedipus' styles (i.e., fall-rise-fall; see Reagan et al., 2016). Comparing these emotional arcs to the number of downloads for each of the fiction titles, the authors suggest that 'Icarus', 'Oedipus', and 'Man in a hole' are the most successful plot shapes.

In a different study IBM's Watson Tone Analyzer was used to examine the continuous trajectories (i.e., in terms of emotion: e.g., joy; language: e.g., analytical language, and personality: e.g., extraversion) in public speaking (Tanveer et al., 2018). The authors investigated the audience's perception of linguistic structures of 2,007 publicly available TED talks. After identifying clusters and comparing these to ratings on the TED website, they found that 'flat' (i.e., less diverse) plot shapes are more likely to be rated as 'long-winded', emphasizing the importance of narrative variety for speech success. Furthermore, the majority of speeches showed a positive ending. While specific linguistic (or visual) characteristics of a video may not be accessible prior to watching a video, its style and structure may play a role in capturing and holding on to the attention of a viewer. Since video streaming services tend to only consider "quality views" (i.e., longer than X amount of time, (YouTube.com, 2018), examining the continuous sentiment styles over narrative time may be a worthwhile endeavor.

## 1.3 The current study

This investigation aims to use intra-textual sentiment analysis to examine the continuous sentiment styles used in YouTube vlogs. Specifically, we primarily aim to identify distinct trajectories of continuous sentiment. In addition, we explore the sentiment styles' relationship to gender and their overall prevalence on YouTube. Besides these aims, we introduce a novel method of extracting sentiment that is sensitive to valence shifters but can be used on non-punctuated data (e.g., YouTube vlogs). We also present a publicly available corpus of YouTube vlog transcripts.

## 2 Data

The data and code used to produce the analysis and data in this paper are publicly available.[2] [3]

### 2.1 Vlog selection

Vlogs were selected by choosing vlog channels (i.e., where YouTubers post their videos) from a list of the top 250 YouTube channels (ranked by subscriber count) in the 'People & Blogs' category (retrieved from www.socialblade.com on 29 March 2018). Channels were selected for inclusion if the videos on the channel were in English and the channel solely focused on vlogs – excluding channels with gaming, music and prank videos. The aim was to select a sample of 20 female and 20 male vloggers from the list of 250 YouTube channels. However, the majority of channels in this list were male or family vloggers, resulting in a somewhat unbalanced sample. After inspecting all 250 channels in the list, 37 channels that satisfied the aforementioned criteria remained.

### 2.2 Scraping vlog transcripts

To obtain vlog transcripts, we accessed all videos published on the selected YouTube channels and used www.downsub.com to access video transcripts. That website downloads and returns YouTube transcripts for a specified video URL directly in the browser. We developed a python script that takes video URLs as input and communicates indirectly with YouTube via downsub.com to request and retrieve video transcripts. We did not differentiate between manually-added and automatically-generated transcripts but encountered several cases in which neither was available for a certain video. In this case, we ignored the affected video and proceeded without considering it further in the analysis.

### 2.3 Preprocessing

The retrieved transcripts were XML-encoded, and the preprocessing included the removal of all XML tags to provide human-readable sequences. Furthermore, each row of the XML transcripts contained the word sequence of the vlog with its corresponding start and end time. Since the YouTube transcripts do not include any punctuation information, we decoded each row by removing the XML tags and merged the resulting human-readable sequences to one continuous string for each vlog.

### 2.4 Feature extraction

The primary aim of this investigation was to examine the continuous sentiment structures of popular vlogs. Taking inspiration from previous work on the narrative arcs of novels (Gao et al., 2016), we wanted to analyze how the sentiment of the spoken content of a vlog moves dynamically throughout the vlog's narrative time. When conducting sentiment extraction, an important consideration is the treatment of valence shifters that influence the meaning of a sentiment. For example, an utterance such as "this was not a bad day" should be rated as a positive sentiment since the valence of "bad" is shifted through the negator "not". Existing methods of extracting sentiment, however, either ignore valence shifters (Jockers, 2015b) or are sentence-based (i.e., they provide weighted sentiments only per sentence, e.g., the *R*-package *sentimentr*, Rinker, 2018b). Since the scraped transcripts are non-punctuated, but we also wanted to account for valence shifters, we built a "naïve context" sentiment extractor that is sensitive to negators (e.g., not, doesn't), [de-]amplifiers (e.g., really, hardly), and adversative conjunctions (e.g., but, however). The rationale is inspired by the algorithm behind *sentimentr* (Rinker, 2018b) but extends that approach to data that are non-punctuated or very brief.

Specifically, our algorithm identifies each sentiment as matched with the 'Jockers & Rinker Polarity Lookup Table' from the *lexicon* R package (Rinker, 2018a) and then constructs a "naïve context" cluster (i.e., without relying on punctuation or other structure) around that sentiment (here: two words before and after the sentiment). For each context cluster, the raw sentiment is then weighted by the presence of valence shifters. For example, "this was not a bad day in the sun" would result in a cluster around the identified sentiment word "bad" of (not a **bad** day in). The weights assigned to the valence shifters in this study (Negator: -1.00, Amplifier: 1.50, De-amplifier: 0.50, Adversative conjunction: 0.25) are motivated by those in the similar software package *sentimentr*.[4] Lastly, the sentiment word in the cluster is replaced by its sentiment value: (-1, 1, -0.75, 1, 1). We can then calculate the product of all vector elements and retrieve the weighted sentiment of that cluster (e.g., 0.75 for "not a bad day in").

---

[2] Data and code for vlog scraping:
https://github.com/ben-aaron188/narrative_structures.

[3] Code for feature extraction algorithm:
https://github.com/ben-aaron188/naive_context_sentiment

We performed the naïve sentiment extraction on each vlog transcript resulting in a vector consisting of zeros (for words that did not match the sentiment lookup table) and weighted sentiment values. That vector was transformed to a standardized narrative time from 0 to 100 using the discrete cosine transformation from the *syuzhet* R package (Jockers, 2015b). The sentiment values were scaled from -1 (lowest sentiment per vlog transcript) to +1 (highest sentiment per vlog transcript). The transformation yielded a vector of 100 sentiment coordinates for each transcript.

## 3 Results

### 3.1 Corpus statistics

The final corpus consisted of the transcripts of 27,333 vlogs from 24 vloggers with a total corpus size of 40,318,924 words and vlogs accounting for more than 24 billion views. These are all vlogs of which we were able to retrieve a transcript of at least ten words. Table 1 shows that female vloggers were underrepresented in the sample. More than half of the vlogs stem from male vloggers and a third from families who vlog. Since view count is highly dependent on the number of days the vlog is online, we corrected the view count for each vlog by dividing it by the number of days it was online. We then excluded 462 vlogs (1.69%) that were considered view count outliers (i.e., more than three standard deviations above the mean). All subsequent analyses were conducted on the final sample with these outliers excluded.

|  | Overall | Female | Male |
|---|---|---|---|
| # of vlog channels | 24 | 12.50% (n = 3) | 54.17% (n = 13) |
| # of videos | 27,333 | 7.64% (n = 2,087) | 49.02% (n = 13,399) |
| Videos/channel | 1138.88 | 695.67 | 949.92 |
| Avg. length (# of words) | 1475.10 (746.53) | 1480.06 (610.72) | 1366.31 (804.68) |
| Avg. view count | 893,932 (2,566,021) | 456,603 (718,895) | 898,089 (1,900,855) |
| Avg. view count (after outlier removal) | 762,143 (1,479,588) | 412,592 (490,792) | 817,029 (1,476,804) |
| Stand. view count | 1553.69 (2733.35) | 1457.40 (2508.95) | 1620.12 (2875.96) |

Table 1. Descriptive statistics (mean, SD).

### 3.2 Identifying continuous sentiment styles

The primary aim of this paper was to examine whether the narrative structure of vlogs can be captured by a few overarching sentiment styles. We used the binned sentiments extracted for each vlog in an unsupervised non-hierarchical *k*-means cluster analysis. Using the within-cluster-sum-of-squares in a scree plot for 1 to 30 clusters, we observed an inflection after seven clusters and therefore decided to build a *k*-means model with *k*=7 (Figure 1 and 2).

The *k*-means model assigned one cluster to each vlog transcript, and we subsequently averaged the continuous sentiment structure of all vlogs belonging to each cluster. Figure 1 and 2 show the distinct average shapes of the continuous sentiment structures of each cluster. The dotted red lines indicate the upper and lower boundaries of the sentiment shape +/- one standard deviation; the blue lines show the 99% confidence intervals. The seven plot shapes represent the average sentiment of all vlogs in that shape cluster in each bin of standardized narrative time (1-100). For example, for the first shape (Figure 1, second plot), the sentiment starts highly positive in the first quantile of the narrative time and then quickly dips below zero to moderately negative sentiment. Halfway through the narrative time, the sentiment neutralizes and becomes negative again until the last quantile from which it gradually becomes less negative and ends in a nearly neutral sentiment.

---
[4] Consensus on values for valence shifter weights has yet to emerge in literature. The algorithm is available as an *R*-implementation to modify these values.

| Cluster | Description | Label | % of vlogs | Avg. length | Stand. view count |
|---|---|---|---|---|---|
| 1 | Highly positive start, followed by semi-negative section, mildly positive section, negative ending | "Downhill from here" | 16.06 | 1477.41 (741.71) | 1524.52 (2738.15) |
| 2 | Negative start, followed by positive section, mildly negative section, semi-positive ending | "Mood swings" | 11.84 | 1503.70 (759.19) | 1504.72 (2702.29) |
| 3 | Negative first half, positive second half | "Rags to riches" | 15.92 | 1454.41 (727.83) | 1557.04 (2697.74) |
| 4 | Positive first half, negative second half | "Riches to rags" | 12.36 | 1485.32 (769.01) | 1477.71 (2615.45) |
| 5 | Semi-positive start, followed by a highly negative section, semi-positive end | "Bump in the road" | 13.46 | 1480.21 (729.47) | 1507.75 (2650.23) |
| 6 | Majority of the narrative is semi-negative, with highly positive end | "End on a high note" | 18.39 | 1507.69 (701.18) | 1818.73 (3030.10) |
| 7 | Majority of the narrative is positive with two peaks | "Twin peaks" | 11.97 | 1402.86 (788.97) | 1368.05 (2511.51) |

Table 2. Narrative styles taxonomy and descriptive statistics.

Table 2 proposes a taxonomy of these seven continuous sentiment styles of YouTube vlogs and provides descriptive statistics for each. We used a linear mixed effects model to test for an effect of the clusters on the view count and transcript length. To account for dependence in the data (i.e., that we have multiple transcripts per vlogger), we included the vlogger as a random effect in the model. The analysis indicated that there was no significant difference in the corrected view count between the clusters; non-standardized $\beta$-coefficient of cluster = 6.78, $se$ = 7.40, $t(451.25)$ = 0.92, $p$ = .360. Nor was there a difference in transcript length; $\beta$ = -1.23, $se$ = 2.07, $t(114.10)$ = -0.60, $p$ = .552. Thus, it is not statistically justified to argue that one particular narrative style attracted more views than others. Nor do these results provide support for a relationship between a vlog's length (as measured by the number of words in its transcript) and a particular sentiment style (e.g., that lengthier vlogs contain more sentiment shifts).

A one-sided Chi-square test revealed that the number of vlogs was not uniform per cluster, $X^2(6)$ = 720.61, $p < .001$. The observed frequencies of vlogs in each cluster deviated significantly from the frequency that is expected if the vlogs were distributed uniformly ($n$ = 3832, 14.29%). Standardized residuals ($z$-scores) of the observed relative to the expected frequencies can be used to examine which observations deviated in which direction. There were significantly more vlogs than expected in the "downhill-from-here" cluster ($z$ = 8.30), the "rags-to-riches" cluster ($z$ = 7.64), and the "end-on-a-high-note" cluster ($z$ = 19.21). The clusters "mood swings" ($z$ = -11.45), "riches-to-rags" ($z$ = -9.02), "bump-in-the-road" ($z$ = -3.86), and "twin peaks" ($z$ = -10.82) were significantly underrepresented. Aside from the considerable overrepresentation of the "end-on-a-high-note" cluster and underrepresentation of the "mood swings" and "twin peaks" clusters, the distribution is rather harmonious.[5]

### 3.3 Additional analysis: Sentiment styles and gender

We also assess whether there is a relationship between gender and the continuous sentiment style. There was significant association as indicated with a 3 (male, female, family) by 7 (clusters) Chi-square test, $X^2(12)$ = 134.82, $p$ < .001. Table 3 shows that family vloggers used the "twin peaks" style significantly more often than expected and used the "end-on-a-high-note" style significantly less often than expected (* = significant at $p$ < .01). For female vloggers, the most used continuous sentiment style was "riches to rags", while they used "end-on-a-high-note" less often than expected. Male vloggers preferred the

| Cluster | Family | Female | Male |
| --- | --- | --- | --- |
| Downhill from here | 2.23 | 1.26 | -2.88* |
| Mood swings | -2.31 | 1.96 | 1.25 |
| Rags to riches | 2.13 | -1.95 | -1.08 |
| Riches to rags | -2.05 | 4.88* | -0.56 |
| Bump in the road | 1.69 | -1.12 | -1.08 |
| End on a high note | -5.16* | -6.03* | 8.32* |
| Twin peaks | 3.83* | 2.25 | -4.99* |

Table 3. Standardized residuals for the cluster-by-gender association.

"end-on-a-high-note" style while they used the styles "downhill from here" and "twin peaks" less often. Since we did not ascertain a balanced gender distribution, these findings should be treated cautiously and subjected to replications on datasets more suitable for gender analysis.

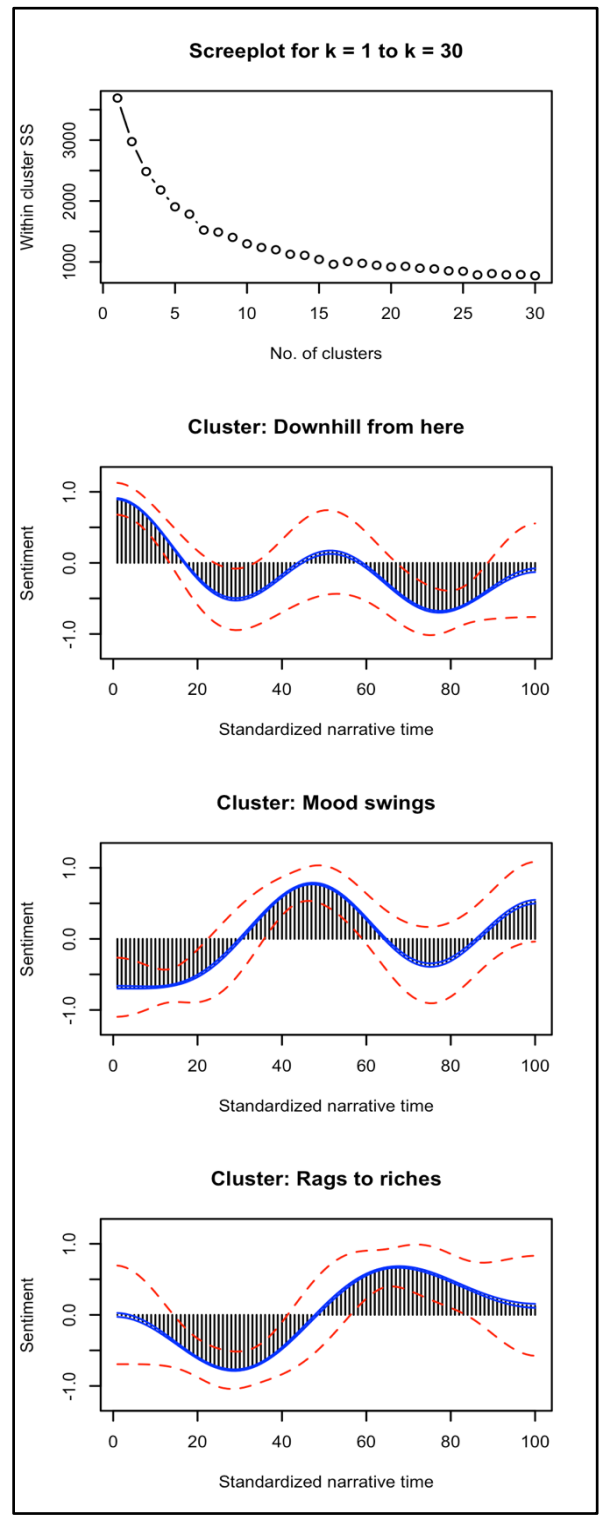

Figure 1. Scree plot and average sentiment style shapes for clusters 1-3. *Note:* Dotted red lines = +/- 1 SD; blue lines = 99% CI.

---

[5] The appendix provides a list of URLs to vlogs and vlog channels typical of a particular style.

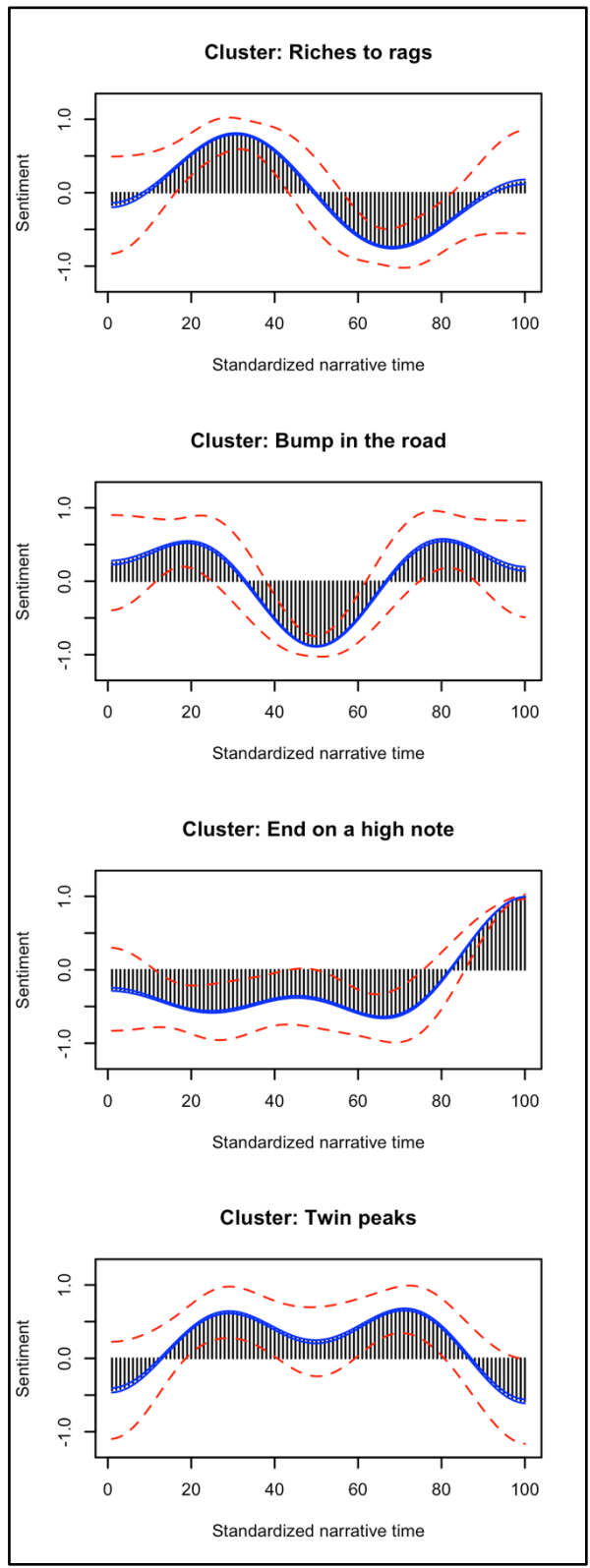

Figure 2. Average sentiment style shapes for clusters 4-7. *Note:* Dotted red lines = +/- 1 SD; blue lines = 99% CI.

## 4 Discussion

In this paper, we examined the continuous sentiment structures of a new corpus of YouTube vlog transcripts using intra-textual sentiment analysis. We were able to identify seven distinct continuous sentiment styles. The "rags to riches", "riches to rags" and the "end-on-a-high-note" styles displayed an alternating pattern with one shift between predominantly positive and negative content. By contrast, the "bump in the road" and "twin peaks" styles show a marked negative middle part of the vlog and an overall positive sentiment with a slightly negative beginning and end, respectively. The most volatile styles were the "downhill from here" and the "mood swings" styles with both showing three shifts in sentiment.

While no continuous sentiment style was related to higher view counts, we observed that the "end-on-a-high-note" style was the most common in our corpus. The pattern of this particular style suggests that vloggers build up their content towards a positive ending following the idea to 'end strong', possibly to engage their viewers to keep watching subsequent videos. A similar trajectory was found for TED speeches (Tanveer et al., 2018), suggesting that YouTube vloggers may employ strategies similar to those of prolific public speakers. However, since the "downhill from here" style is second most frequent in our dataset, these findings do not indicate any tendency to a general sentiment development over time. In fact, the "downhill from here" and "end on a high note" styles exhibit a somewhat opposing behavior since the first gradually decreases the sentiment over time whereas the second aims at developing a positive sentiment towards the end of the video. When projecting this observation to the gender-specific findings, one can see that the differences between family, female and male vloggers might account for these contradictory findings. Both family and female vloggers (representing 45.83% of all vloggers) used the "downhill from here" more often than male vloggers, whereas male vloggers used the "end on a high note" style most often.

Gender differences might help further understand vlog style choice: family vloggers used the "twin peaks" style more often than female or male vloggers, with the latter significantly under-using that style. Female vloggers resorted to the "riches to rags" style most often, while male vloggers preferred the "end-on-a-high-note" style.

Vlogs made by females started with a definite positive buildup followed by a marked dip towards clear negative sentiment and a gradual increase towards the positive near the end of the vlog. Family vloggers were the only ones who preferred a style that ended with a sentiment leading toward the negative.

Contrary to the findings of Tanveer et al. (2018), we did not find support for the notion that diverse styles characterized by higher sentiment volatility are more popular: both YouTuber's themselves, as well as the audience, did not show a preference for the most volatile narrative styles. The intra-textual analysis approach helped us to uncover dynamics in the sentiment of vlogs that would otherwise have been blurred by an overall sentiment score. In many cases (e.g., for the "rags to riches" and "riches to rags" styles) the sentiment dynamics would have canceled each other out to a neutral sentiment.

### 4.1 Limitations

Despite the interesting initial findings using the new corpus of vlog transcripts and the linguistic-temporal analysis, the current study is not without its limitations. First, we partially relied on automated transcription of the vlogs, so the retrieved transcripts consisted of texts generated with YouTube's speech recognition software. Although recent advancements in machine learning research provide promising improvements to speech recognition (LeCun et al., 2015; Zhang et al., 2017b; Zhang et al., 2017a) and contributed to an accuracy increase of fifty percent for YouTube's automatic captions in English (Official YouTube Blog, 2017), the automatic generation of YouTube video captions remains a challenging task (Liao et al., 2013). It could be that potential inaccuracies might have affected our findings and future studies could set out to replicate our results using user-provided high-quality vlog transcripts.

Second, the sample consisted of YouTubers that already are successful concerning subscriber count, and we might, therefore, have painted a skewed picture of the vlogging landscape. For future studies, it would be interesting to look at potential moderating variables such as the vloggers 'vlogging age' or the vlog's topic. Possibly, the use of sentiment (and other linguistic dimensions) might differ between starting YouTubers and those who are more prolific and regularly attract millions of views.

Third, by using vlog transcripts, we resorted solely to the linguistic modality of vlogs and did not look at the use of visual aspects in the vlogs. Using a similar dynamic method, future studies could look at how the use of visual content behaves and interacts with language use over the narrative time.

### 4.2 Future work

Using the corpus, method, and findings presented here as a starting point, we hope that future research can extend this exploratory work.

- For example, it would be interesting to perform unsupervised clustering analyses on the vlog level to find "vlog-twins" or even "vlogger-twins". Using observations that are similar in a multidimensional feature space could be exploited for semi-experimental studies where the effects of individual variables (e.g., a change in vlogging style) can be isolated towards drawing conclusions of causal nature.

- Moreover, the sentiment is only one dimension of the linguistic aspects of vlogs, and the current analysis might be extended to different constructs. For example, a significant challenge lies in detecting extremist content on social media platforms like YouTube (Burgess, 2017) and the dynamic approach might help uncover extremist parts in videos and enable a more fine-grained inspection of vlogs.

- The temporal aspect of narrative time could be broadened so that the evolution of the narrative style across videos of individual vloggers can be captured. This would be a useful method to identify changes in vlogging strategy.

- Finally, the intra-textual method might help in detecting online misinformation or deception within texts. A major challenge, for example, for linguistic deception detection lies in identifying lies embedded within a mostly truthful statement (Bachenko et al., 2008). Similarly, it would be interesting to examine whether a dynamic linguistic-temporal approach as used here can aid in the detection of misinformation online

(e.g., Pérez-Rosas et al., 2018). The method used here and in a few related studies (Jockers, 2015a; Gao et al., 2016) could, therefore, also be of interest to researchers across disciplines.

## 5 Conclusion

Vlogging is a unique and novel means of communication. We explored the transcripts of vlogs as a source for linguistic analysis, and, by looking at intra-textual sentiment dynamics of each vlog, we identified seven distinct continuous sentiment styles. Vlogs ending on a positive note were the most prevalent, and we observed that gender was associated with different vlogging style preferences. The current paper presented an initial glimpse at the rich data source of vlogs, and a dynamic sentiment analysis approach helped uncover continuous sentiment structures. As such we hope the corpus, method, and findings presented here function as an impetus towards more analyses on this emerging means of online communication.

## References

Oya Aran, Joan-Isaac Biel, and Daniel Gatica-Perez. 2014. Broadcasting oneself: Visual discovery of vlogging styles. *IEEE Transactions on multimedia*, 16(1):201–215.

Joan Bachenko, Eileen Fitzpatrick, and Michael Schonwetter. 2008. Verification and implementation of language-based deception indicators in civil and criminal narratives. In *Proceedings of the 22nd International Conference on Computational Linguistics-Volume 1*, pages 41–48. Association for Computational Linguistics.

Joan-Isaac Biel and Daniel Gatica-Perez. 2010. Vlogcast yourself: Nonverbal behavior and attention in social media. In *International Conference on Multimodal Interfaces and the Workshop on Machine Learning for Multimodal Interaction*, page 50. ACM.

Youmna Borghol, Sebastien Ardon, Niklas Carlsson, Derek Eager, and Anirban Mahanti. 2012. The untold story of the clones: content-agnostic factors that impact YouTube video popularity. In *Proceedings of the 18th ACM SIGKDD international conference on Knowledge discovery and data mining*, pages 1186–1194. ACM.

Matt Burgess. 2017. Google's using a combination of AI and humans to remove extremist videos from YouTube. http://www.wired.co.uk/article/google-youtube-ai-extremist-content

Erik Cambria. 2016. Affective computing and sentiment analysis. *IEEE Intelligent Systems*, 31(2):102–107.

Tommaso Caselli and Piek Vossen. 2016. The Storyline Annotation and Representation Scheme (StaR): A Proposal. In *Proceedings of the 2nd Workshop on Computing News Storylines (CNS 2016)*, pages 67–72, Austin, Texas. Association for Computational Linguistics.

Jianbo Gao, Matthew L. Jockers, John Laudun, and Timothy Tangherlini. 2016. A multiscale theory for the dynamical evolution of sentiment in novels. In *Behavioral, Economic and Socio-cultural Computing (BESC), 2016 International Conference on*, pages 1–4. IEEE.

Matthew Jockers. 2015a. Revealing Sentiment and Plot Arcs with the Syuzhet Package. http://www.matthewjockers.net/2015/02/02/syuzhet/

Matthew Jockers. 2015b. *Syuzhet: Extract Sentiment and Plot Arcs from Text*. https://github.com/mjockers/syuzhet

Bennett Kleinberg, Yaloe van der Toolen, Aldert Vrij, Arnoud Arntz, and Bruno Verschuere. 2018. Automated verbal credibility assessment of intentions: The model statement technique and predictive modeling. *Applied Cognitive Psychology*, 32(3):354–366, May.

Yann LeCun, Yoshua Bengio, and Geoffrey Hinton. 2015. Deep learning. *Nature*, 521(7553):436.

Hank Liao, Erik McDermott, and Andrew Senior. 2013. Large scale deep neural network acoustic modeling with semi-supervised training data for YouTube video transcription. In *Automatic Speech Recognition and Understanding (ASRU), 2013 IEEE Workshop on*, pages 368–373. IEEE.

Rada Mihalcea and Carlo Strapparava. 2009. The lie detector: Explorations in the automatic recognition of deceptive language. In *Proceedings of the ACL-IJCNLP 2009 Conference Short Papers*, pages 309–312. Association for Computational Linguistics.

Heather Molyneaux, Susan O'Donnell, Kerri Gibson, and Janice Singer. 2008. Exploring the gender divide on YouTube: An analysis of the creation and reception of vlogs. *American Communication Journal*, 10(2):1–14.

Official YouTube Blog. 2017. One billion captioned videos. https://youtube.googleblog.com/2017/02/one-billion-captioned-videos.html

Myle Ott, Claire Cardie, and Jeffrey T. Hancock. 2013. Negative deceptive opinion spam. In *HLT-NAACL*, pages 497–501.

Veronica Pérez-Rosas, Bennett Kleinberg, Alexandra Lefevre, and Rada Mihalcea. 2018. Automatic detection of fake news. In *Proceedings of the 27th International Conference on Computational Linguistics, COLING 2018*, Santa Fe, New Mexico, USA, August.

Andrew J. Reagan, Lewis Mitchell, Dilan Kiley, Christopher M. Danforth, and Peter Sheridan Dodds. 2016. The emotional arcs of stories are dominated by six basic shapes. *EPJ Data Science*, 5(1):31.

Tyler Rinker. 2018a. *lexicon: Lexicon Data*. http://github.com/trinker/lexicon

Tyler Rinker. 2018b. *sentimentr: Calculate Text Polarity Sentiment*. http://github.com/trinker/sentimentr

M. Iftekhar Tanveer, Samiha Samrose, Raiyan Abdul Baten, and M. Ehsan Hoque. 2018. Awe the Audience: How the Narrative Trajectories Affect Audience Perception in Public Speaking. In *Proceedings of the 2018 CHI Conference on Human Factors in Computing Systems*, page 24. ACM.

James Thorne, Andreas Vlachos, Christos Christodoulopoulos, and Arpit Mittal. 2018. FEVER: a large-scale dataset for Fact Extraction and VERification. *arXiv:1803.05355 [cs]*, March. arXiv: 1803.05355.

YouTube.com. 2018. How video views are counted - YouTube Help. https://support.google.com/youtube/answer/2991785?hl=en

Ying Zhang, Mohammad Pezeshki, Philémon Brakel, Saizheng Zhang, Cesar Laurent Yoshua Bengio, and Aaron Courville. 2017a. Towards end-to-end speech recognition with deep convolutional neural networks. *arXiv preprint arXiv:1701.02720*.

Yu Zhang, William Chan, and Navdeep Jaitly. 2017b. Very deep convolutional networks for end-to-end speech recognition. In *Acoustics, Speech and Signal Processing (ICASSP), 2017 IEEE International Conference on*, pages 4845–4849. IEEE.

Renjie Zhou, Samamon Khemmarat, and Lixin Gao. 2010. The impact of YouTube recommendation system on video views. In *Proceedings of the 10th ACM SIGCOMM conference on Internet measurement*, pages 404–410. ACM.